\documentclass[twoside,11pt]{article}
\usepackage{jmlr2e}

\usepackage{algorithm}
\usepackage{algorithmic}
\usepackage{wrapfig}
\usepackage{graphicx} 

\DeclareGraphicsExtensions{.pdf,.jpeg,.png}

\usepackage{wrapfig}
\usepackage{subfiles}
\usepackage[subpreambles=false]{standalone}

\usepackage{microtype}
\usepackage{graphicx}
\usepackage{subcaption}
\usepackage{booktabs} 
\usepackage{url}
\usepackage{graphicx}
\usepackage{amsmath}
\usepackage{amssymb}
\usepackage{wrapfig}
\usepackage{color}

\usepackage{enumitem}

\usepackage{color}

\newcommand*\diff{\mathop{}\!\mathrm{d}}

\newcommand{\R}{{\mathbb R}}   
\newcommand{\E}{{\mathbb E}}   

\newcommand{\eat}[1]{}

\newcommand{\T}[1]{{\mathcal{#1}}} 
\newcommand{\V}[1]{{\mathbf{#1}}} 

\newcommand{\refn}[1]{(\ref{#1})}

\newcommand{\brck}[1]{\left(#1\right)}

\newcommand{\brckcur}[1]{\left\{#1\right\}}

\newcommand{\fr}[2]{\frac{#1}{#2}}

\newcommand{\be}{\begin{equation}}
\newcommand{\ee}{\end{equation}}
\newcommand{\bali}{\begin{eqnarray*}}
\newcommand{\eali}{\end{eqnarray*}}
\newcommand{\eq}[1]{\begin{align}#1\end{align}}

\newcommand{\calI}{\mathcal{I}}

\newcommand{\mathR}{\mathbb{R}}

\newcommand{\ti}[1]{\textit{#1}}

\newcommand{\trnn}{\texttt{HOT-RNN}}
\newcommand{\tlstm}{HOT-LSTM}

\usepackage{hyperref}

\jmlrheading{1}{2019}{1-48}{4/00}{10/00}{yu18}{Rose Yu, Stephan Zheng, Anima Anandkumar, Yisong Yue}

\ShortHeadings{Higher-Order Tensor RNNs}{Yu et al.}
\firstpageno{1}

\begin{document}

\title{Long-Term Forecasting using  Higher-Order Tensor RNNs}

\author{\name Rose Yu \email rose@caltech.edu 
       \AND
       \name Stephan Zheng \email stephan@caltech.edu 
       \AND
      \name Anima  Anandkumar \email anima@caltech.edu 
       \AND
       \name Yisong Yue \email yyue@caltech.edu \\
        \addr Department of Computing and Mathematical Sciences\\
       California Institute of Technology\\
       Pasadena, CA 91125, USA}

\editor{Francis Bach, David Blei and Bernhard Sch{\"o}lkopf}

\maketitle

\begin{abstract}
  We present Higher-Order Tensor RNN (\trnn{}), a novel family of neural sequence architectures for multivariate forecasting in environments with nonlinear dynamics.
  Long-term forecasting in such systems is highly challenging, since there exist long-term temporal dependencies, higher-order correlations and sensitivity to error propagation.
  Our proposed recurrent architecture addresses these issues by learning the nonlinear dynamics directly using higher-order moments and higher-order state transition functions.
  Furthermore, we decompose the higher-order structure using the tensor-train decomposition to reduce the number of parameters while preserving the model performance.
  We theoretically establish the approximation guarantees and the variance bound for \trnn{} for general sequence inputs. We also
  demonstrate $5 \sim 12\%$  improvements for long-term prediction over general RNN and LSTM architectures on a range of simulated environments with nonlinear dynamics, as well on real-world time series data.
\end{abstract}


\section{Introduction}
\label{intro}

\begin{wrapfigure}{r}{0.36\textwidth}
\vspace{-18pt}
\begin{center}
\includegraphics[width=\linewidth]{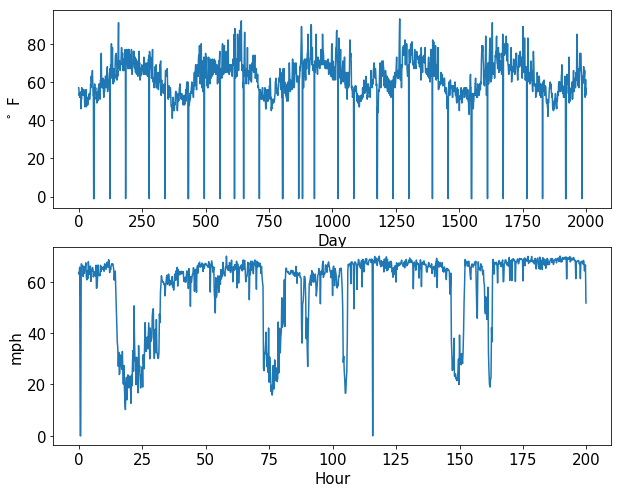}
\caption{Climate and traffic time series visualization. The time series can be viewed as a realization of highly nonlinear dynamics. 
}
\label{fig:time_series}
\vspace{-20pt}
\end{center}
\end{wrapfigure}

One of the central questions in science is forecasting: given the past history, how well can we predict the future?
In many domains with complex multi-variate correlation structures and nonlinear dynamics, forecasting is highly challenging since the system has long-term temporal dependencies and higher-order dynamics. Examples of such systems abound in science and engineering, from biological neural network activity, fluid turbulence, to climate and traffic systems (see, e.g., Figure \ref{fig:time_series}).
Since current forecasting systems are unable to faithfully represent the higher-order dynamics, they have limited ability for accurate \ti{long-term} forecasting.

Therefore, a fundamental challenge is accurately modeling nonlinear dynamics and obtaining stable long-term predictions, given a dataset of realizations of the dynamics.
Here, the forecasting problem can be stated as follows: how can we efficiently learn a model that, given only a few initial states, can  predict a sequence of future states over a long horizon of $T$ time-steps accurately and reliably?

Common approaches to forecasting include classic linear time series models such as auto-regressive moving average (ARMA), state space models such as hidden Markov model (HMM), and deep neural networks. See a survey on time series forecasting by \citep{box2015time} and the references therein.
A recurrent neural network (RNN), as well as its memory-based extensions such as the LSTM, is a class of models that have achieved state of the art performance on sequence prediction tasks from demand forecasting \citep{ flunkert2017deepar} to speech recognition \citep{soltau2016neural} and video analysis \citep{lecun2015deep}.
But most of these methods focus on short-term predictions, and often fail to generalize to nonlinear dynamics and forecast over long time horizons.

In this work, we propose \trnn{}, a model class that is more expressive and empirically generalizes better than standard RNNs, for the same model capacity.
%
%
\trnn{} explicitly models the 1) \ti{higher-order dynamics}, by
incorporating a longer history  and higher-order state interactions of previous hidden states; and
2) using \ti{tensor trains decomposition} that greatly reduces the number of model parameters, while mostly preserving the correlation structure of the full-rank model. 
We prove that
\trnn{} is exponentially more expressive than standard RNNs for functions that satisfy certain regularity conditions. Our contributions can be summarized as follows:
\begin{itemize} 
\item We propose a novel family of RNNs \trnn{}s to encode non-Markovian dynamics and higher-order state interactions. To address the memory issue, we propose a tensor-train decomposition that makes learning tractable.
\item We provide theoretical guarantees for the expressiveness of \trnn{}s for nonlinear dynamics, and characterize the target dynamics and its \trnn{} representation. In contrast,  no such theoretical results are known for standard recurrent networks.
\item We show that \trnn{}s can forecast more accurately for significantly longer time horizons  compared to standard RNNs and LSTMs on simulated data and real-world environments with nonlinear dynamics.
\end{itemize}

\section{Related Work}
\label{related}
\paragraph{Time series forecasting}
Time series forecasting is at the core of many dynamics modeling tasks. In statistics, classic work  such as the ARMA or ARIMA model \citep{box2015time}  model a stochastic process with assumptions of linear dynamics. In the control and dynamical system community, estimating dynamics  models from measurement data is also known as \ti{system identification} \citep{ljung2001system}.  System identification often requires strong parametric assumptions which are often challenging to find from first principles.  Moreover, finding (approximate) solutions of complex nonlinear differential equations demands high computational cost. In our work, we instead take a ``mode-free'' approach to learn a powerful approximate nonlinear dynamics model.

\paragraph{Recurrent Neural Networks}
Using neural networks to model time series data has a long history  \citep{schmidhuber2015deep}. Recent developments in deep leaning and RNNs has led to non-linear forecasting models such as deep AutoRegressive  \citep{flunkert2017deepar},  Predictive State Representation \citep{downey2017practical}, Deep State Space model \citep{rangapuram2018deep}. However, these works usually study short-term forecasting and use RNNs that contain only the most recent state.  Our method contrasts with this by explicitly modeling higher-order dyanmics to capture long-term dependencies.

There are several classic work on higher-order RNNs. For example, \citep{giles1989higher} proposes a higher-order RNN to simulate a deterministic finite state machine and recognize regular grammars. The model considers a multiplicative structure of inputs and the most recent hidden state, but is limited to two-way interactions. \citep{sutskever2011generating} also studies tensor RNNs that allow a different hidden-to-hidden weight matrix for every input dimension.
\cite{soltani2016higher} proposes a higher-order RNN that concatenates a sequence of past hidden states, but the underlying state interactions are still linear. 
Moreover, hierarchical RNNs \citep{zheng2016generating} have been used to model sequential data at multiple temporal resolutions. Our method generalizes all these works to capture higher-order interactions using a hidden-to-hidden tensor.

\paragraph{Tensor methods}
Tensor methods  have tight connections with neural networks. For example,  \citep{novikov2015tensorizing, stoudenmire2016supervised} employ tensor-train to compress the weights in neural networks.  \citep{yang2017tensor}  extends this idea to RNNs by reshaping  the inputs  into a  tensor and factorizes the input-hidden weight tensor. However, the purpose of these works is model compression in the input space whereas our method learns the dynamics in the hidden state space. Theoretically, \citep{cohen2016expressive} shows convolutional neural networks and  hierarchical tensor factorizations are equivalent. \citep{khrulkov2017expressive} provides expressiveness analysis for shallow networks using tensor train.

Tensor methods have also been used for sequence modeling. For example, one can apply tensor decomposition as  method of moments estimators for latent variable models such as Hidden Markov Models (HMMs)  \citep{anandkumar2012method}. Tensor methods have also shown promises in reducing the model dimensionality of multivariate spatiotemporal learning problems \citep{yu2016learning}, as well as  nonlinear system identification \citep{decuyper2019decoupling}. Most recently, \cite{schlag2018learning} combine tensor product of relational information and recurrent neural networks for natural language reasoning tasks.  This work however, to the best of our knowledge, is the first to consider tensor networks within RNNs  for sequence learning in environments with nonlinear dynamics.


%
\section{Higher-Order Tensor RNNs}
\label{trnn}
\paragraph{Forecasting Nonlinear Dynamics}
Our goal is to learn an efficient forecasting model for \ti{continuous multivariate time series} in environments with nonlinear dynamics.
The state $\V{x}_t \in \mathR^d$ of such systems evolves over time using a set of \ti{nonlinear} differential equations:
\eq{\label{eq:dynamics}
\brckcur{\xi^i\brck{\V{x}_t, \fr{d\V{x}}{dt}, \fr{d^2\V{x}}{dt^2}, \ldots; \phi} = 0 }_i,
}
where $\xi^i$ can be an arbitrary (smooth) function of the state $\V{x}_t$ and its derivatives. 
Continuous time dynamics are usually described by differential equations while  difference equations are employed for discrete time. 
In continuous time, a classic example is the first-order Lorenz attractor, whose realizations showcase the ``butterfly-effect'', a characteristic set of double-spiral orbits. 
In discrete-time, a non-trivial example is the 1-dimensional Genz dynamics, whose difference equation is:
\eq{\label{eq:genzprodpeak}
	x_{t+1} = \brck{c^{-2} + (x_t + w)^2}^{-1}, \hspace{10pt}  c,w \in [0,1],
}
where $x_t$ denotes the system state at time $t$ and $c,w$ are the parameters. Due to the nonlinear nature of the dynamics, such systems exhibit higher-order correlations, long-term dependencies and sensitivity to error propagation, and thus form a challenging setting for forecasting.
%

Given a sequence of initial states $\V{x}_0\ldots \V{x}_t$, the forecasting problem aims to learn a dynamics model $F$ that outputs a sequence of future states $\V{x}_{t+1} \ldots \V{x}_T$. 
\eq{\label{eq:forecast}
F: \brck{\V{x}_0\ldots \V{x}_t} \mapsto \brck{\V{y}_{t} \ldots \V{y}_T},
\hspace{10pt} \V{y}_t = \V{x}_{t+1},
}
The system is governed by some unknown dynamics. Hence, accurately approximating the dynamics is critical to learning a good forecasting model and making predictions for long time horizons.

\begin{figure*}[t]
\begin{center}
\begin{minipage}[t]{0.62\linewidth}
\centering		\includegraphics[width=\linewidth]{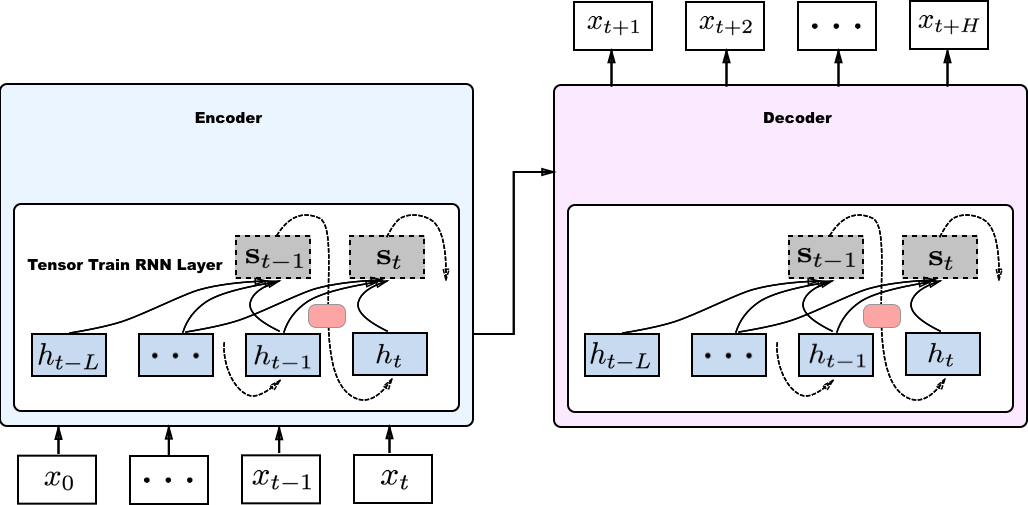}
\caption{\trnn{} within a seq2seq model. Both encoder  and decoder contain higher-order recurrent cells. The augmented state $\V{s}_{t-1}$ (grey) takes in past $L$ hidden states (blue) and forms a higher-order tensor. \trnn{} (red)  factorizes the tensor and outputs the next hidden state.}
\label{fig:seq2seq}
\end{minipage}
\hspace{0.02\linewidth}
\begin{minipage}[t]{0.33\linewidth}
\centering		\includegraphics[width=\linewidth]{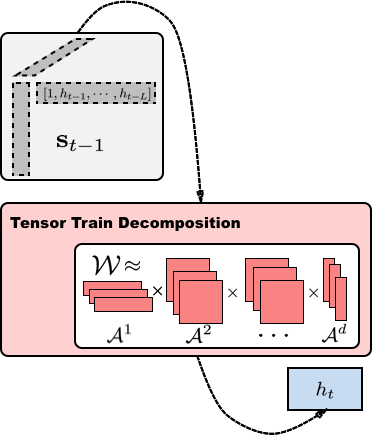}
\caption{A \trnn{} cell. The augmented state $\V{s}_{t-1}$ (grey) forms a higher-order tensor, which is then factorized to output the next hidden state.}
\label{fig:ttrnn}
\end{minipage}
\end{center}
\vspace{-5mm}
\end{figure*}

\paragraph{First-order Markovian Models}
In deep learning, popular approaches such as recurrent neural networks (RNNs) employ first-order hidden-state models to approximate the dynamics. An RNN with a single  cell recursively  computes a hidden state $\V{h}_t$ using the most recent hidden state $\V{h}_{t-1}$, generating  the output $\V{y}_t$ from the hidden state $\V{h}_t$ :
\eq{\label{eq:rnn}
\V{h}_t = f(\V{x}_t, \V{h}_{t-1}; \theta_f),\hspace{10pt} \V{y}_t = g(\V{h}_t; \theta_g),
}
where $f$ is the state transition function, $g$ is the output  function and $\{\theta_f, \theta_g\}$ are the corresponding  model parameters. A common parametrization scheme for \refn{eq:rnn} applies a nonlinear  activation function such as sigmoid $\sigma$ to a linear map of $\V{x}_t$ and $\V{h}_{t-1}$ as:
\eq{
\V{h}_t &= \sigma(W^{hx} \V{x}_t + W^{hh} \V{h}_{t-1} + \V{b}^h), \quad
\V{x}_{t+1} = W^{xh} \V{h}_t + \V{b}^x,
}
where $W^{hx}, W^{xh}$ and $W^{hh}$ are  the transition weight matrices and $\V{b}^h, \V{b}^x$ are the biases.

RNNs have many different variations, including LSTMs \citep{hochreiter1997long} and GRUs \citep{chung2014empirical}. 
%
Although a RNN can approximate any function in theory, its hidden state $\V{h}_t$  only depends on the previous state $\V{h}_{t-1}$ and the input $\V{x}_t$. Such models do not explicitly capture higher-order dynamics and only  implicitly encode long-term dependencies between all historical states $\V{h}_{0} \ldots \V{h}_{t}$. This limits the representation power of RNNs, especially for forecasting in environments with nonlinear dynamics. Hence, instead of using a wide RNN with many hidden units, we exploit the recurrent cell to design higher-order tensor RNNs that can approximate complex non-linear governing equations.



\subsection{Higher-Order Non-Markovian Models}

To effectively learn nonlinear dynamics with higher-order temporal dependency, we propose a family of models that generalizes standard RNNs: higher-order recurrent neural networks, or  \trnn{}. 
We design \trnn{}s with two goals in mind: explicitly modeling 1) $L$-order Markov processes with $L$ steps of temporal memory and 2) polynomial interactions between the hidden states $\V{h}_{\cdot}$ and $\V{x}_t$.

First, we consider longer ``history'': we keep length $L$ historic states: $\V{h}_{t},\cdots, \V{h}_{t-L}$:
\eq{
\V{h}_t = f( \V{x}_t , \V{h}_{t-1}, \cdots, \V{h}_{t-L}; \theta_f)
\label{eqn:high_order_markov}
}
where $f$ represents the state transition function.  In principle, early work \citep{giles1989higher} has shown that with a large enough hidden state size, such recurrent structures are capable of approximating any dynamical system.

Second, to learn the nonlinear dynamics $\xi$ efficiently, we also use higher-order moments to approximate the state transition function.
We use an augmented state $\V{s}$, where we mute the subscript of $\V{s}_{t-1}$ for notation simplicity.:
\begin{equation}
	\V{s}^T = [1 \hspace{5pt} \V{h}_{t-1}^\top \hspace{5pt} \ldots \hspace{5pt} \V{h}_{t-L}^\top ]
\end{equation}
which concatenates $L$ previous hidden states.
To compute $\V{h}_t$, we construct a $P$-dimensional transition \ti{weight tensor} to model degree-$P$ polynomial interactions between hidden states:
%
\begin{align}
[\V{h}_{t}]_\alpha = \phi(W^{hx}_\alpha\V{x}_t+  
    \sum_{i_1,\cdots, i_p}\T{W}_{\alpha i_1 \cdots i_{P}}  \underbrace{\V{s}_{i_1} \otimes\cdots\otimes \V{s}_{i_p} }_{P} )\nonumber
\label{eqn:tensor_rnn}
\end{align}
where $\alpha$ indices the hidden dimension, $i_\cdot$ indices the higher-order terms and $P$ is the total  polynomial order. We included the bias unit $1$ in $\V{s}$ to account for the first order term, so that  $\V{s}_{i_1} \otimes\cdots\otimes \V{s}_{i_p} = [1, \V{h}_t, \V{h}_t\V{h}_{t-1},\cdots]$ can include all  polynomial expansions of hidden states up to order $P$. 

%
%
%
The \trnn{} with LSTM cell, or ``\tlstm{}'', is defined analogously as:
\begin{align}
[\V{i}_t, \V{g}_t, &\V{f}_t, \V{o}_t]_\alpha = \sigma (W^{hx}_\alpha \V{x}_t + \sum_{i_1,\cdots, i_p}\T{W}_{\alpha i_1 \cdots i_{P}}  \underbrace{\V{s}_{i_1} \otimes\cdots\otimes \V{s}_{i_P} }_{P} ), \\
& \V{c}_t = \V{c}_{t-1} \circ \V{f}_t +  \V{i}_t\circ \V{g}_t,
\qquad
\V{h}_t = \V{c}_t \circ \V{o}_t \nonumber
\end{align}
where $\circ$ denotes the Hadamard product. Note that the bias units are again included.

 \trnn{} is a basic  unit that can be incorporated in most of the existing recurrent neural architectures such as convolutional RNN \citep{xingjian2015convolutional} and hierarchical RNN \citep{chung2016hierarchical}. In this work, we use  \trnn{} as a module for sequence-to-sequence (seq2seq) framework \citep{sutskever2014sequence} in order to perform long-term forecasting.

As shown in Figure \ref{fig:seq2seq}, seq2seq models consist of an encoder-decoder pair. The encoder takes an input sequence and learns a hidden representation. The decoder initializes with this hidden representation and generates an output sequence. Both the encoder and the decoder contain multiple layers of higher-order tensor recurrent cells (red).
The augmented state $\V{s}_{t-1}$ (grey) concatenates the past $L$ hidden states;
the \trnn{} cell takes $\V{s}_{t-1}$ and outputs the next hidden state. 
The encoder encodes the initial states $x_{0}, \ldots, x_{t}$ and the decoder predicts $x_{t+1}, \ldots, x_{T}$. 
For each time step $t$, the decoder uses its previous prediction $\V{y}_t$ as an input.
\subsection{Dimension Reduction with Tensor-Train}
Unfortunately, due to the ``curse of dimensionality'', the number of parameters in $\T{W}_\alpha$ with hidden size $H$ grows exponentially as $O(HL^P)$, which makes the higher-order model prohibitively large to train. To overcome this difficulty, we  utilize   \textit{tensor networks} to approximate the weight tensor. Such networks encode a structural decomposition of tensors into low-dimensional components and have been shown to provide the most general approximation to smooth tensors \citep{orus2014practical}.
The most commonly used tensor networks are \textit{linear tensor networks} (LTN), also known as \textit{tensor-trains} in numerical analysis or \textit{matrix-product states} in quantum physics \citep{oseledets2011tensor}.

A tensor train model decomposes a $P$-dimensional tensor $\T{W}$ into a network of sparsely connected low-dimensional tensors $\{\T{A}^p \in \R^{r_{p-1} \times n_p \times r_{p}} \}$ as:
\begin{equation*}
\T{W}_{i_1 \cdots i_P} =
\sum_{\alpha_1 \cdots \alpha_{P-1}}
\T{A}^1_{\alpha_0 i_1 \alpha_1}%
\T{A}^2_{\alpha_1 i_2 \alpha_2}%
\cdots%
\T{A}^P_{\alpha_{P-1} i_P \alpha_P}
\nonumber
\end{equation*}
 with $ \alpha_0 = \alpha_P = 1$, as depicted in Figure (\ref{fig:ttrnn}). When $r_0 = r_{P} = 1$ the $\{r_p\}$ are called the tensor-train rank.
With tensor-train decomposition, we can reduce the number of parameters of \trnn{} from $(HL+1)^{P}$ to $(HL+1)R^2P$, with $R = \max_p{r_p}$ as the upper bound on the tensor-train rank.
Thus, a major benefit of tensor-train is that they \textit{do not} suffer from the curse of dimensionality, which is in sharp contrast to many classical tensor decomposition models, such as the Tucker decomposition.


%
\section{Approximation Theorem for HOT-RNNs}
\label{thm}
%
A significant benefit of using \trnn{} is that we can theoretically characterize its expressiveness  for approximating the underlying dynamics.   The main idea is to analyze a class of functions that satisfies certain regularity conditions. For such functions, tensor-train representations preserve the weak differentiability and yield a compact representation.

The following theorem characterizes the representation power of \trnn{}, viewed as a one-layer hidden neural network, in terms of 1) the regularity of the target function $f$, 2) the dimension of the input space, 3) the tensor train rank and 4) the order of the tensor:
\begin{theorem}
Let the target function $f\in \mathcal{H}^k_\mu$ be a H\"older continuous function defined on a input  domain $\mathcal{I} =I_1\times \cdots \times I_d$, with  bounded derivatives up to order $k$ and finite Fourier magnitude distribution $C_f$. A single layer \trnn{} with $h$ hidden units, $\hat{f}$ can approximate $f$ with approximation error $\epsilon$ at most:
\eq{
\epsilon \leq \fr{1}{h} \brck{C_f^2 \frac{d-1}{(k-1)(r+1)^{k-1}} + C(k)p^{-k} }
}
where $C_f = \int |\omega|_1 |\hat{f}(\omega) d \omega|$, $d$ is the dimension of the function, i.e., the size of the state space, $r$ is the tensor-train rank, $p$ is the degree of the higher-order polynomials i.e., the order of the tensor, and $C(k)$ is the coefficient of the spectral expansion of $f$.
\label{eqn:thm}
\end{theorem}

\textbf{Remarks}: The result above shows that the number of weights required to approximate the target function $f$ is dictated by its regularity (i.e., its H\"older-continuity order $k$). The expressiveness of \trnn{} is driven by the selection of the rank $r$ and the polynomial degree $p$; moreover, it improves for functions with increasing regularity.  Compared with ``first-order'' regular RNNs, \trnn{}s are  exponentially more powerful for large rank: if the order  $p$ increases, we require fewer hidden units $h$. 
%
%

\ti{Proof sketch}: 
For the full proof, see the Appendix. 
We design \trnn{} to approximate the underlying system dynamics. The target function $f(\V{x})$ represents the state transition function, as in \refn{eqn:tensor_rnn}.
We first show that if $f$ preserves weak derivatives, then it has a compact tensor-train representation. Formally, let us assume that $f$ is a Sobolev function: $f\in\mathcal{H}^k_\mu$, defined on the input space $\T{I}= I_1 \times I_2\times \cdots I_d $, where each $I_i$ is a set of vectors. The space $\mathcal{H}^k_\mu$ is defined as the functions that have bounded derivatives up to some order $k$ and are $L_\mu$-integrable.
\begin{eqnarray}
\mathcal{H}^k_\mu =  \left\{  f  \in L_\mu(\T{I}):\sum_{i\leq k}\|D^{(i)}f\|^2   < +\infty \right\},
\end{eqnarray}
where $D^{(i)}f$ is the $i$-th weak derivative of $f$ and $\mu \geq 0$.\footnote{A weak derivative generalizes the derivative concept for (non)-differentiable functions and is implicitly defined as: e.g. $v\in L^1([a,b])$ is a weak derivative of $u\in L^1([a,b])$ if for all smooth $\varphi$ with $\varphi(a) = \varphi(b) = 0$: $\int_a^bu(t)\varphi'(t) = -\int_a^bv(t)\varphi(t)$.}
%
It is known that any Sobolev function admits a Schmidt decomposition: $f(\cdot) = \sum_{i =0}^\infty \sqrt{\lambda_i } \gamma (\cdot)_i \otimes \phi (\cdot)_i $, where $\{\lambda \}$ are the eigenvalues and $\{\gamma\}, \{ \phi\}$ are the associated eigenfunctions.
Hence, for $\V{x}\in\calI$, we can represent the target function $f(\V{x}) $ as an infinite summation of products of a set of basis functions:
\begin{align}
&f(\V{x}) = \sum_{\alpha_0,\cdots,\alpha_d=1}^\infty
%
\T{A}^1 (x_1)_{\alpha_0 \alpha_1}
\cdots
\T{A}^d(x_d)_{\alpha_{d-1}	 \alpha_d},
\label{eqn:ftt}
\end{align}
where  $ \{ \T{A}^j(x_j)_{\alpha_{j-1} \alpha_j} \}$ are basis functions over each input dimension.
These basis functions satisfy $\langle \T{A}^j(\cdot)_{im}, \T{A}^j (\cdot)_{in} \rangle = \delta_{mn}$ for all $j$.
If we truncate \eqref{eqn:ftt} to a low dimensional subspace ($\V{r}<\infty$), we obtain a functional approximation of the state transition function $f(\V{x})$.  
This approximation is also known as the \ti{functional tensor-train} (FTT):
\begin{align}
&f_{FTT}(\V{x}) = \sum_{\alpha_0,\cdots,\alpha_d}^\mathbf{r}
\T{A}^1(x_1)_{\alpha_0\alpha_1}
\cdots
\T{A}^d(x_d)_{\alpha_{d-1}\alpha_d},
\end{align}

In practice, \trnn{} implements a polynomial expansion of the states using $[\V{s}, \V{s}^{\otimes 2}, \cdots, \V{s}^{\otimes P}]$, where $P$ is the degree of the polynomial. The final function represented by \trnn{} is a polynomial approximation of the functional tensor-train function $f_{FTT}$. 

Given a target  function $f(\V{x}) = f(\V{s}\otimes \dots \otimes \V{s})$, we can express it using FTT and the polynomial expansion of the states $\V{s}$. This allows us to characterize \trnn{} using a family of functions that it can represent.  Combined with the classic neural network approximation theory \cite{barron1993universal}, we can bound the approximation error for \trnn{} with one hidden layer. The above results applies to the full family of \trnn{}s, including those using  vanilla RNN or LSTM as the recurrent cell.

One can think of the universal approximation result in Theorem \ref{eqn:thm} bounds the estimation bias of the model: $f-\mathbb{E} [\hat{f}]$, where the expectation is taken over training sets. While a large neural network can approximate any function, training as a large neural network will be hard given a finite data set, demonstrating bias-variance trade-off. In the next section, we provide bounds for the estimation variance.

\section{Variance Bound for \trnn{}}
\label{gen}

Given a time series from a $P$-th order dynamics $(X_1, \cdots, X_P)$,  denote the variance over the joint hidden states as  $\hat{C} := \sum_{i=1}^m \otimes [\T{A}(X^{(i)}_1), \cdots ,\T{A}(X^{(i)}_P)] $, where $\T{A}(\cdot)$ are the basis functions. Define the variance of the estimated dynamics as $C := \E_{X_1 X_2 \cdots,  X_P}[\phi(X_1)\otimes \phi(X_2), \cdots, \otimes \phi(X_P)]$, with $\phi(\cdot)$ being the feature mapping. The following theorem bounds the estimation variance of the \trnn{}. 
\begin{theorem}[Estimation Variance Bound]
Assuming the time series is governed by a system whose order of dynamics is at most $P$, represented as a joint  probabilistic distribution $P(X_1,\cdots, X_P)$.  The variance for the  \trnn{} estimator $\hat{C}$ and the true variance $C$ of  the population statistics is  upper bounded by:
\[   \|\hat{C} - C \| = \mathcal{O}_p \big( \frac{2\sqrt{2} \rho^{P/2}\sqrt{\log \frac{2}{\delta}}}{\sqrt{m}} \big) \]

where $C := \E_{X_1 X_2 \cdots,  X_P}[\phi(X_1)\otimes \phi(X_2), \cdots, \otimes \phi(X_P)]$ and $m$ is number of samples, $\rho:=\sup\limits_{x\in \T{X}}k(x,x)$.
\end{theorem}

\begin{figure}[htbp]
	\begin{center}
		\begin{subfigure}[b]{0.55\linewidth}		\includegraphics[width=0.9\linewidth]{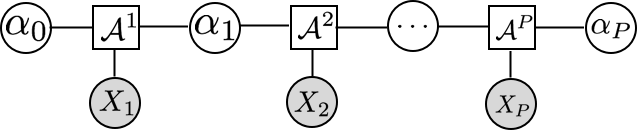}
			\caption{Tensor Train Graph}
			\label{fig:lorenz_data}
		\end{subfigure}
		\begin{subfigure}[b]{0.4\textwidth}		\includegraphics[width=0.9\linewidth]{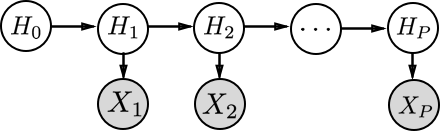}
			\caption{Conditional Random Field}
			\label{fig:lorenz_20}
		\end{subfigure}
	\end{center}
\vskip -0.1in
\caption{The graphical representation of tensor train model and conditional random field. The circles denote hidden variables and the shaded circles represent observed variables.}
\label{fig:tt-crf}
\end{figure}

\ti{Proof }: 
The tensor-train distribution forms a Gibbs field, thus based on Hammersley-Clifford Theorem, a Gibbs field satisfies global Markov property, therefore, it must be a  Conditional Random Field (CRF) as the  conditional probability distribution factorizes. According to the duality between tensor network and graphical model \cite{robeva2017duality}, tensor train is the dual graph of CRF. Figure \ref{fig:tt-crf} visualizes such dual relationship in the graphical model template. 
After establishing the relationship between conditional random field and functional tensor-train, we can bound the variance of the \trnn{} estimator. 

We first state the well-celebrated Hammersley-Clifford theorem that gives the sufficient and necessary conditions of which a probability distribution  is a Markov Random field.

\begin{theorem}[Hammersley-Clifford]
A  graphical model  $G$ is a \textit{Markov Random Field} if and only if
the probability distribution $P(X)$ on $G$ is a \textit{Gibbs distribution}:
\[P(X)= \frac{1}{Z} \prod_{c\in C_G}\psi(X_c) \]
where $Z=\sum_x  \prod_{c\in C_G}\psi(X_c)$ is the normalization constant, $\psi$ are functions defined on maximal cliques, and $C_G$ is a set of all maximal cliques in the graph.
\label{thm:hc}
\end{theorem}

When the underlying distribution is a Markov random  field and belongs to the exponential family, we can generalize Theorem \ref{thm:hc} to kernel functions and obtain the following results.
\begin{lemma}[Kernelized Hammersley-Clifford] \cite{altun2012exponential}
Given a Markov random field $X$ with respect to a graphical model $G$, if  the sufficient statistics $\Phi(X) = (\Phi(X_{c_1}) , \cdots, \Phi(X_{c_i}) )$, then the kernels  $k(X, X')=\langle \Phi(X), \Phi(X')\rangle $ satisfy
\[k(X, X') = \sum_{c\in C_G}k_c(X_c, X_c')\]
where  $\Phi(X_c)$ are the sufficient statistics defined on maximal cliques, $k_c(X, X') = \langle \Phi(X_C), \Phi(X'_C) \rangle $
\label{thm:khc}
\end{lemma}


Our tensor train model factorizes the state transition function  
$\V{h}_t = \V{x}_t+  f(\V{s}_1 \otimes \cdots \otimes \V{s}_p)$, where each augmented hidden states $\V{s} =  [1, \V{h}_{t-1}, \cdots, \V{h}_{t-L}]$. 
For a joint distribution of $d$ variables  $X_1, \cdots, X_P$, taking values from $1$ to $n$ from a conditional random field. Their joint probability density table is a d-dimensional tensor $P(X_1, \cdots, X_P) \in \R^{n^P}$, with the values of the variable act as indices. Following the result  of Lemma \ref{thm:khc},  the feature  functions  $\T{A}(X_p)$ factorize over maximum cliques:
\begin{eqnarray*}
P(X_1,\cdots, X_P)&=& \sum_{\alpha_0, \cdots, \alpha_P}P(X_1,\cdots,X_d,\alpha_0,\cdots, \alpha_P)\\
&=& \sum_{\alpha_0,\cdots,\alpha_P=1}^\infty
%
\T{A}^1 (X_1)_{\alpha_0 \alpha_1}
\cdots
\T{A}^d(X_d)_{\alpha_{P-1}	 \alpha_P},
\end{eqnarray*}
which is the functional tensor-train model.


Consider a time series of dynamics up to $P$-th order,  we can view the time series as a joint distribution over  $P$ variables  $P(X_1, X_2, \cdots, X_P)$.   The covariance of such joint distribution is defined as  $C := \E_{X_1 X_2 \cdots,  X_P}[\phi(X_1)\otimes \phi(X_2), \cdots, \otimes \phi(X_P)]$. \trnn{} estimates the joint distribution by factorizing the  tensor product of the feature mapping functions: $\hat{C} := \sum_{i=1}^m \otimes [\T{A}(X^{(i)}_1), \cdots ,\T{A}(X^{(i)}_P)] $.   Given $m$ i.i.d. samples of times series, $\mathcal{D} = \{{X}_1^{(i)}, {X}_2^{(i)}, \cdots, {X}_P^{(i)} \}_{i=1}^m$, we can then generalize the results from \cite{song2013nonparametric} for multi-view latent variable models to \trnn{}.




%
\section{Experiments}
\label{exp}
We conducted  exhaustive experiments to examine the behavior of the proposed \trnn{} model on both synthetic and real-world time series data. The source code is available at \url{https://github.com/yuqirose/tensor_train_RNN}.
 
\subsection{Datasets}
We validated the accuracy and efficiency of \trnn{} on the following three datasets. 
\paragraph{Genz}
Genz functions are often used as  basis for evaluating high-dimensional function approximation. In particular, they have been used to analyze tensor-train decompositions \citep{bigoni2016spectral}. There are in total $7$ different Genz functions. (1) $g_1 (x) = \cos( 2 \pi w + cx)$, (2) $g_2 (x) =( c^{-2} + (x+w)^{-2})^{-1}$, (3) $g_3(x) = (1+cx)^{-2}$, (4) $e^{- c^2\pi (x-w)^2}$ (5)  $e^{-c^2 \pi | x-w|}$ (6) $g_6(x) = \begin{cases} 0  \quad x>w\\ e^{cx} \quad  else \end{cases}$.  For each function, we generated a dataset with $10,000$ samples using \refn{eq:genzprodpeak} with $w=0.5$ and $c=1.0$ and random initial points draw from a range of $[-0.1, 0.1]$.

\paragraph{Traffic}
We use the traffic data of Los Angeles County highway network collected from  California department of transportation \footnote{\url{http://pems.dot.ca.gov/}}. The dataset consists of $4$ month speed readings aggregated every $5$ minutes . Due to large number of missing values ($\sim 30\%$) in the raw data, we  impute the missing values using the average values of non-missing entries from other sensors at the same time. In total, after processing, the dataset covers $35\,136,$ time-series. We treat each sequence as daily traffic of $288$ time stamps. We up-sample the dataset every $20$ minutes, which results in a dataset of $8\,784$ sequences of daily measurements. We select $15$ sensors as a joint forecasting tasks.

\paragraph{Climate}
We use the daily maximum temperature data from the U.S. Historical Climatology Network (USHCN) daily  \footnote{\url{http://cdiac.ornl.gov/ftp/ushcn\_daily/}}. The dataset contains daily measurements for $5$ climate variables for approximately  $124$ years. The records were collected across more than $1\,200$ locations and span over $45\,384$ days. We analyze the area in California which contains $54$ stations. We removed the first $10$ years of day, most of which has no observations. We treat the temperature reading per year as one sequence and impute the missing observations using other non-missing entries from other stations across years. We augment the datasets by rotating the sequence every $7$ days, which results in a data set of $5\,928$ sequences.

%

 \begin{figure*}[t]
 	\vskip 0.2in
 \begin{subfigure}[b]{0.32\linewidth}
 	\centering
 	\includegraphics[width=\linewidth]{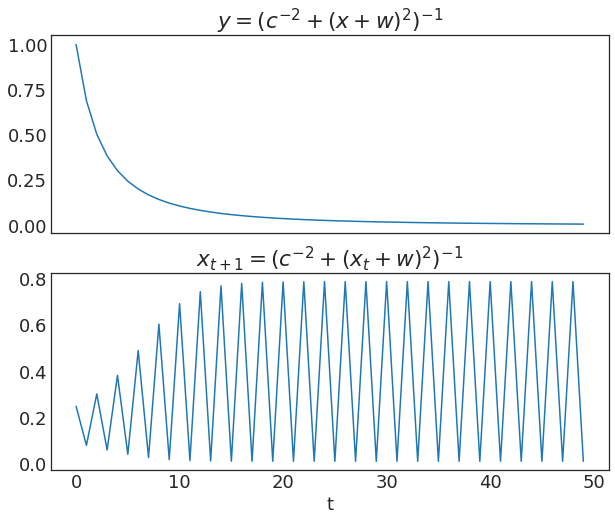}
 	\caption{\textsl{Genz dynamics}}
 	\label{fig:genz}
 \end{subfigure}
     \begin{subfigure}[b]{0.32\linewidth}
         \centering
         \includegraphics[width=\linewidth]{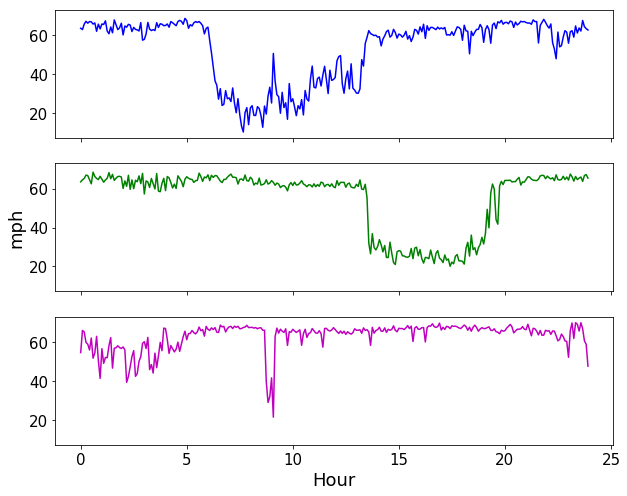}
         \caption{\textsl{Traffic} daily : 3 sensors  }
                 \label{fig:traffic}
     \end{subfigure}
     \begin{subfigure}[b]{0.32\linewidth}
         \centering
         \includegraphics[width=\linewidth]{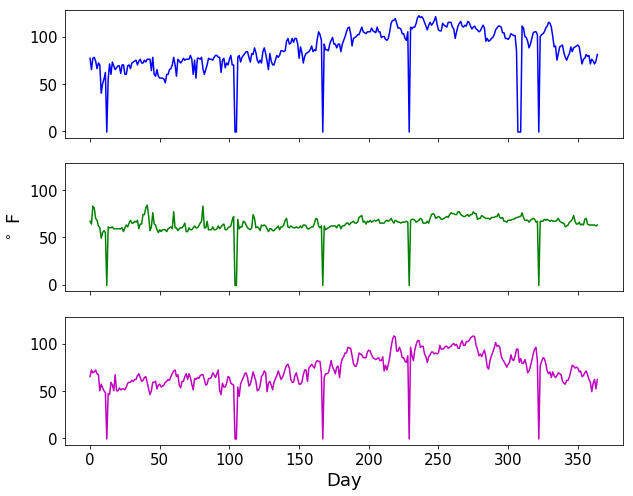}
         \caption{\textsl{Climate} yearly: 3 stations }
                 \label{fig:climate}
     \end{subfigure}
     \caption{Data visualizations:
     (\ref{fig:genz}) Genz dynamics,
     (\ref{fig:traffic}) traffic data,
     (\ref{fig:climate}) climate data.}
     \label{fig:data_visual}
     \vskip -0.2in
 \end{figure*}

Figure \ref{fig:data_visual} visualizes the time series from Genz dynamics, traffic and climate systems, respectively.  To test the stationarity of the time series, we also perform a Dickey–Fuller test on the real-world traffic and climate data. Dickey–Fuller  test is a commonly used statistical test procedure  to determine whether a time series is stationary. Its null hypothesis is that  a unit root is present in an autoregressive model, hence the time series  is not stationary.  The test statistics of the traffic and climate data is shown in Table \ref{app:tb:adf}, which demonstrate the non-stationarity of the time series.
\begin{table}[hb]
    \begin{center}
        \begin{tabular}{c |  cc | cc }
        \hline
           & \multicolumn{2}{c}{\bf Traffic } & \multicolumn{2}{c}{\bf   Climate } \\ 
           \hline
    Test Statistic      & 0.00003 & 0   & 3e-7 & 0  \\
p-value        &     0.96& 0.96  &     1.12 e-13 & 2.52 e-7  \\
Number Lags Used  &     2 & 7  &  0 & 1 \\
Critical Value (1\%)     &   -3.49& -3.51     &   -3.63& 2.7     \\
Critical Value (5\%)  & -2.89& -2.90   & -2.91& -3.70\\
Critical Value (10\%)       & -2.58 & -2.59   & -2.60 & -2.63\\
\hline
        \end{tabular}
    \caption{ Dickey-Fuller test  statistics for traffic and climate data used in the experiments. }
	\label{app:tb:adf}
    \end{center}
\vspace{-5mm}
\end{table}

\subsection{Training Details}

\paragraph{Setup}
We use a seq2seq architecture with \trnn{} using LSTM as recurrent cells (\tlstm{}).
For all experiments, we use the length-$T$ sequence regression loss
%
$L(y,\hat{y}) = \sum_{t=1}^T ||\hat{y}_t-y_t||^2_2,$
%
where $y_t = x_{t+1}, \hat{y}_t$ are the ground truth and model prediction respectively.  For all datasets, we used a $80\%-10\%-10\%$ train-validation-test split and train for a maximum of $1e^4$ steps. We compute the moving average of  the validation loss as an early stopping criteria. We also did not include scheduled sampling \cite{bengio2015scheduled}, as we found training with scheduled sampling became highly unstable under a range of annealing schedules.

\paragraph{Hyperparameter Search}
All models are trained using  RMS-prop with a learning rate decay of $0.8$.  We performed an exhaustive search over the hyper-parameters for validation.  Table \ref{app:tb:hyper} reports the  search range of different hyper-parameters used in this work.

\begin{figure}[t]
\begin{center}
    \begin{subfigure}[t]{0.32\linewidth}
    		\centering
    		\includegraphics[width=\linewidth]{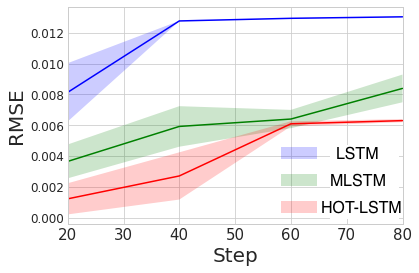}
    		\caption{\textsl{Genz dynamics}}
    \end{subfigure}
   \begin{subfigure}[t]{0.32\linewidth}
        \centering
		\includegraphics[width=\linewidth]{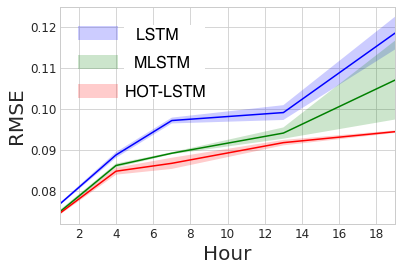}
		\caption{\textsl{Traffic}}
    \end{subfigure}
    \begin{subfigure}[t]{0.32\linewidth}
	\centering
	\includegraphics[width=1.05\linewidth]{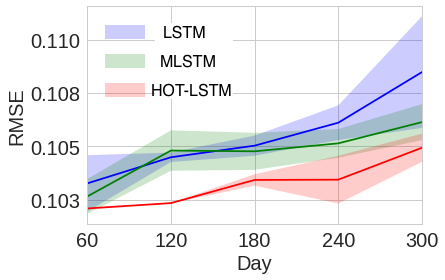}
	\caption{\textsl{Climate}}
\end{subfigure}
    \caption{
    Long-term forecasting RMSE for \ti{Genz dynamics}  and real world \textsl{traffic}, \textsl{climate} time series (best viewed in color).  Comparison of LSTM, MLSTM, and \tlstm{} for varying forecasting horizons given same initial inputs. Results are averaged over $3$ runs. 
    }
    \label{fig:error_horizon}
\end{center}
     	\vskip -0.2in
\end{figure}
\begin{table}[htpb]
\centering
    \begin{tabular}{cc | cc}
    \hline
            \multicolumn{4}{c}{\bf Hyper-parameter search range} \\
            \hline 
            learning rate    & $10^{-1}\ldots 10^{-5}$   &hidden state size  &$8, 16, 32, 64, 128$  \\
        tensor-train rank  &$1 \ldots 16$ & number of lags &$1 \ldots 6$ \\
        number of orders & $1 \ldots 3$  &  number of layers  & $1 \ldots 3$ \\
        \hline
        \end{tabular}
    \caption{Hyper-parameter search range statistics for \trnn{} experiments.}
	\label{app:tb:hyper}
\end{table}

\paragraph{Baselines}
We compared \trnn{} against 2 sets of natural baselines: 1st-order RNN (vanilla RNN, LSTM), and matrix RNNs (vanilla MRNN, MLSTM), which use matrix products of multiple hidden states without factorization \citep{soltani2016higher}.
We observed that \trnn{} with RNN cells outperforms vanilla RNN and MRNN, but using LSTM cells performs best in all experiments. We also evaluated the classic ARIMA time series model  with AR lags of $1 \sim 5$, and MA lags of $1 \sim 3$.   We  observed that it consistently performs $\sim 5\%$ worse than LSTM.
\subsection{Long-term Forecasting Accuracy}
We evaluate the long-term forecasting accuracy of the proposed method and the baselines. 
For \textsl{traffic}, we forecast up to $18$ hours ahead with $5$ hours as  inputs.
For \textsl{climate}, we forecast up to $300$ days ahead given $60$ days of  observations.
For \textsl{Genz dynamics}, we forecast for $80$ steps given $5$ initial steps.
We report the forecasting results  averaged over $3$ runs.

\begin{wraptable}{r}{.5\linewidth}
\vspace{-22pt}
\begin{center}
\resizebox{\linewidth}{!}{%
\begin{sc}
\begin{tabular}{lccc}
\multicolumn{4}{c}{\bf Moving-MNIST (RMSE $\times 10^{-2}$)} \\
& LSTM & MLSTM & \tlstm{} \\
\midrule
$T=20 $ & 9.45 & 9.92 & \textbf{8.94}\\
$T=40 $ & 10.04 & 9.94 & \textbf{9.92}
\end{tabular}
\end{sc}
}
\caption{Sequence-averaged per-pixel RMSE on Moving MNIST.}
\label{tab:movingmnist}
\end{center}
\vspace{-5mm}
\end{wraptable}
Figure \ref{fig:error_horizon} shows the test prediction error (in RMSE) for varying forecasting horizons for different datasets.
We can see that \tlstm{} notably outperforms all baselines on all datasets in this setting. In particular, \tlstm{} is more robust to long-term error propagation.
We observe two salient benefits of using \trnn{}s over the unfactorized models.
First, MRNN and MLSTM can suffer from overfitting as the number of weights increases.
Second, on \textsl{traffic}, unfactorized models also show considerable instability in their long-term predictions.
These results suggest that \trnn{}s learn more stable representations that generalize better for long-term horizons.
To compare the performance on high-dimensional time series, we also evaluated on the unsupervised video prediction task for \textsl{Moving MNIST}. We forecast $20$ and $40$ frames ahead given $10$ initial frames. The per-pixel forecasting RMSE results are shown in Table \ref{tab:movingmnist}. We observe a small gain ($\sim 2-5\%$) of \tlstm{} over the baselines. This is likely due to the fact that the underlying circular dynamics are still pretty simple. Moreover, the high-dimensional inputs have spatial structure that are hard to learn by RNNs alone (note we do not use convolutional features). We expect \tlstm{} to improve over baselines even more with more complicated dynamics and using convolutional features.

%

\begin{figure}[ht]
        \vskip 0.2in
    \begin{center}
    \begin{subfigure}[b]{0.31\textwidth}
        \includegraphics[width=\linewidth]{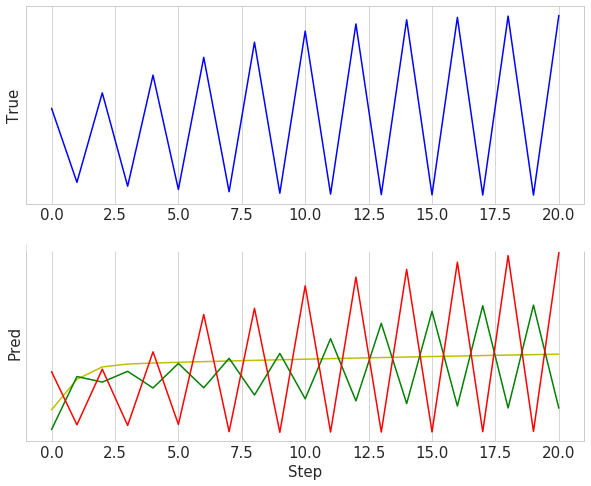}
        \label{fig:f2}
    \end{subfigure}
    ~ 
    \begin{subfigure}[b]{0.31\textwidth}
        \includegraphics[width=\linewidth]{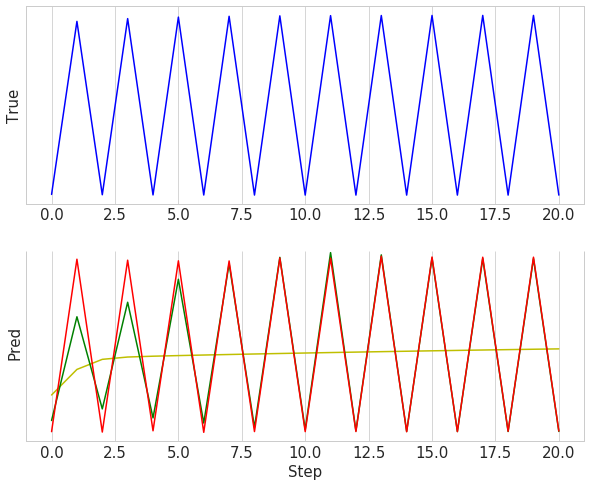}
        \label{fig:df2}
    \end{subfigure}
    ~ 
    \begin{subfigure}[b]{0.31\textwidth}
        \includegraphics[width=\linewidth]{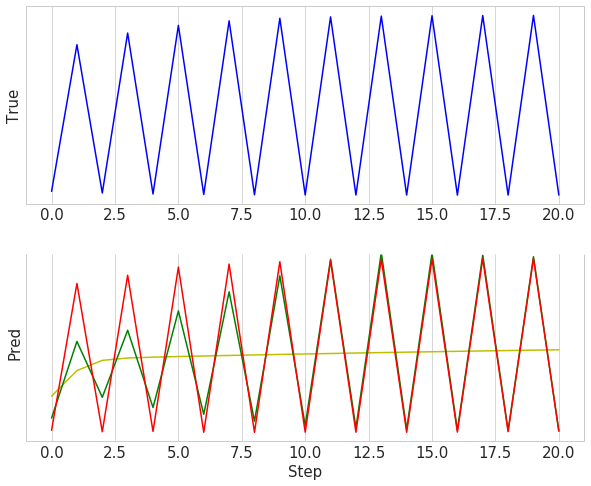}
        \label{fig:df2}
    \end{subfigure}
    \caption{Model prediction for three Genz dynamics ``product peak'' with different initial conditions. Top (blue): ground truth. Bottom: model predictions for LSTM (green) and \tlstm{} (red). \tlstm{} perfectly captures the Genz oscillations, whereas the LSTM fails to do so (left) or only approaches the ground truth towards the end (middle and right).}
        \label{fig:genz_pred}
\end{center}
        \vskip -0.2in
\end{figure}

\begin{figure}[ht]
\begin{center}
    \begin{subfigure}[b]{0.3\textwidth}
        \includegraphics[width=\textwidth]{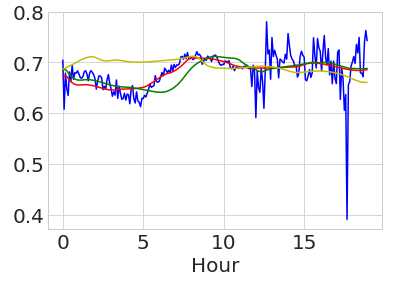}
    \end{subfigure}
    \begin{subfigure}[b]{0.32\textwidth}
        \includegraphics[width=\textwidth]{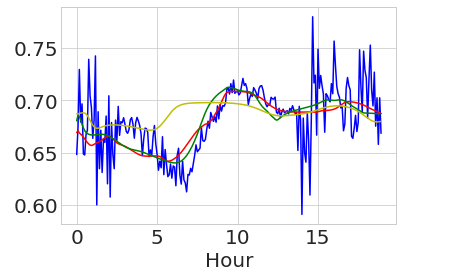}
    \end{subfigure}
    \begin{subfigure}[b]{0.3\textwidth}
        \includegraphics[width=\textwidth]{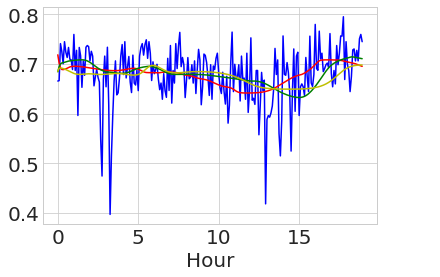}
    \end{subfigure}\\
    \begin{subfigure}[b]{0.32\textwidth}
        \includegraphics[width=\textwidth]{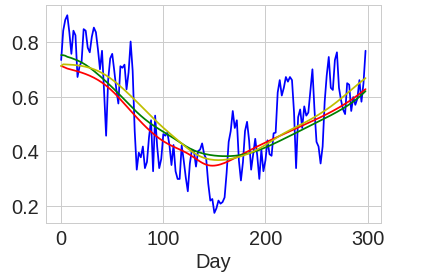}
    \end{subfigure}
    \begin{subfigure}[b]{0.32\textwidth}
        \includegraphics[width=\textwidth]{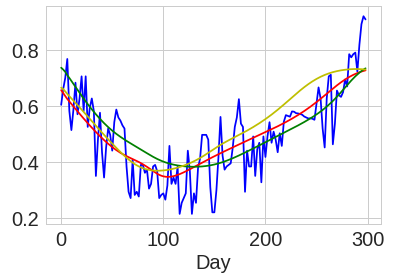}
    \end{subfigure}
    \begin{subfigure}[b]{0.32\textwidth}
        \includegraphics[width=\textwidth]{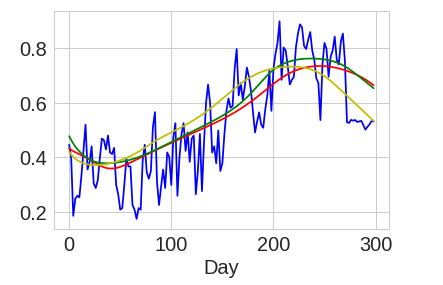}
    \end{subfigure}
    \caption{Top: $18$ hour ahead predictions for hourly \textsl{traffic} time series  given $5$ hour as input for LSTM, MLSTM and \tlstm{}.
    Bottom: $300$ days ahead predictions for daily \textsl{climate} time series given $2$ month observations as input for LSTM, MLSTM and \tlstm{}.}
    \label{fig:pred_vis}
\end{center}
        \vskip -0.2in
\end{figure}
\subsection{Visualization of Predictions}
To get intuition for the learned models, we visualize predictions from the best performing \tlstm{} and baselines.  Figure \ref{fig:genz_pred}  shows the  predictions for the Genz function ``corner-peak'' as the state-transition function from three realizations of Genz dynamics. We can see that \tlstm{} can almost perfectly recover the original function, while LSTM and MLSTM only correctly predict the mean. These baselines cannot capture the dynamics fully, often predicting an incorrect range and phase for the dynamics.

Figure \ref{fig:pred_vis} shows predictions for the traffic and climate datasets. This work uses deterministic models, hence the predictions correspond to the trend.
We can see that the \tlstm{} aligns significantly better with ground truth in long-term forecasting.
As the ground truth time series is highly nonlinear and noisy, LSTM often deviates from the general trend.
While both MLSTM and \tlstm{} can correctly learn the trend, \tlstm{} captures more detailed curvatures due to higher-order structure.

%
%


\subsection{Model Capacity} 
The number of parameters for \trnn{} is $\mathcal{O} (HL+1)R^2P$ with hidden size $H$, lag $L$, rank $R$ and order $P$. This gives us more flexibility to decide the model capacity. Fewer parameters may have limited representation power, while more parameters would cause overfitting. 
\begin{wraptable}{r}{0.35\linewidth}
\caption{Best performing model size on traffic and climate data.}
\begin{tabular}{ccc}
\multicolumn{3}{c}{\bf  Number of Parameters}\\
TLSTM & MLSTM & LSTM \\
\midrule
7,200  & 9,700  & 8,700
\end{tabular}
\label{tb:bestmodel}
\end{wraptable}
Note that the memory complexity only grows quadratically with the rank $R$ while Theorem \ref{eqn:thm} shows the expressiveness of \trnn{} improves exponentially.
In practice, we used cross-validation to select the values for these hyper-parameters. The best models on real-world climate and traffic data are listed in Table \ref{tb:bestmodel}. We can see that the number of parameters of \tlstm{} model is comparable with that of MLSTM and LSTM. 


\begin{figure*}[htbp]
\begin{center}
    \begin{minipage}[b]{0.48\linewidth}

    \begin{center}
                 \includegraphics[width=\linewidth]{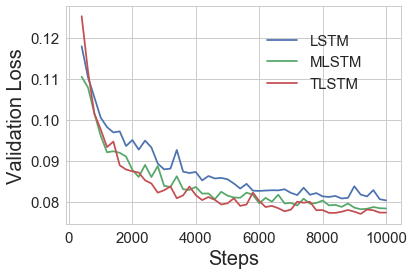}
         \caption{Training speed evaluation of different models: validation loss versus number of steps. Results  are reported using the models with the best long-term forecasting accuracy.}
        \label{fig:ts_speed}      
   
    \end{center}
   \end{minipage}
    \hspace{0.01\linewidth}
   \begin{minipage}[b]{0.48\linewidth}
   \begin{center}
   \begin{sc}
\resizebox{\linewidth}{!}{%
    \begin{tabular}{lcccc}
    \multicolumn{5}{c}{\bf \tlstm{} Prediction Error (RMSE $\times 10^{-2}$)} \\    
    Rank $r$  & 2 & 4 & 8 & 16 \\   	    
    \midrule
    Genz ($T=95$)     & \bf{0.82}  & 0.93  & 1.01    & 1.01  \\
    Traffic ($T=67$)  & 9.17  & \bf{9.11}  & 9.32 & 9.31 \\
    Climate ($T=360$)  & 10.55 & \bf{10.25} & 10.51 & 10.63
    \end{tabular}
	}
    \end{sc}
    \captionof{table}{\tlstm{} performance for varying tensor rank $r$ with $L=3$.}
    \label{tb:hyperpara}
   
   \begin{sc}
   \resizebox{\linewidth}{!}{%
       \begin{tabular}{lcccc}
        \multicolumn{5}{c}{\bf \tlstm{} Traffic Prediction Error (RMSE $\times 10^{-2}$)} \\    
    Lags $L$  & 2 &  4 &5 & 6\\
   	\midrule
    $T=12 $   & \bf{7.38}   &	7.41&7.43&	7.41 \\
    $T=84 $   & \bf{8.97}	  & 9.31 &	9.38&	9.01 \\
    $T=156 $  &9.49   &	9.32	&9.48& \bf{9.31} \\
    $T=228 $  &10.19  & 9.63&	\bf{9.58} &	9.94
    \end{tabular}
    }
       \end{sc}
    \captionof{table}{\tlstm{} performance for various lags $L$ and prediction horizons $T$.}
    \end{center}
    \end{minipage}
\end{center}
\vskip -0.25in
\end{figure*}

\subsection{Speed Performance Trade-off}
 We now investigate potential trade-offs between accuracy and computation.
 Figure \ref{fig:ts_speed} displays the validation loss with respect to the number of steps, for the best performing models on long-term forecasting. We see that \trnn{}s converge faster than other models, and achieve lower validation-loss.
 This suggests that \trnn{} has a more efficient representation of the nonlinear dynamics, and can learn much faster as a result.

\subsection{Sensitivity Analysis}
%
The \tlstm{} model has several hyperparameters, such as tensor-train rank and lag $L$. 
We study the sensitivity of \tlstm{} to these hyperparameters; Table \ref{tb:hyperpara} shows the results. 
In the top row, we report the prediction RMSE for the largest forecasting horizon  w.r.t tensor ranks for all the datasets with lag $3$. 
When the rank is too low, the model does not have enough capacity to capture non-linear dynamics. 
When the rank is too high, the model starts to overfit. 
In the bottom row, we report the effect of changing lag $L$. For each setting, the best $r$ is determined by cross-validation. 
Note that the best lag $L$ also varies for different forecasting horizons.

\subsection{Chaotic Nonlinear Dynamics}

Chaotic dynamics such as Lorenz attractor is notoriously different to lean in non-linear dynamics. In such systems, the dynamics are highly sensitive to perturbations in the input state: two close points can move exponentially far apart under the dynamics.  We  also evaluated tensor-train neural networks on long-term forecasting for Lorenz attractor and report the results.

\paragraph{Lorenz}
The Lorenz attractor system describes a two-dimensional flow of fluids:
\begin{eqnarray*}
	\frac{\diff x}{\diff t} = \sigma(y-x), \hspace{10pt}
	\frac{\diff y}{\diff t} = x(\rho -z) - y, \hspace{10pt}
	\frac{\diff z}{\diff t} = xy - \beta z, \hspace{10pt}
	\sigma=10, \rho=28, \beta=2.667.
\end{eqnarray*}
This system has chaotic solutions (for certain parameter values) that revolve around the so-called Lorenz attractor.
We simulated $10\,000$ trajectories with the discretized time interval length $0.01$. We sample from each trajectory every $10$ units in Euclidean distance. 
\begin{wrapfigure}{r}{0.4\linewidth}
	\includegraphics[width=0.9\linewidth]{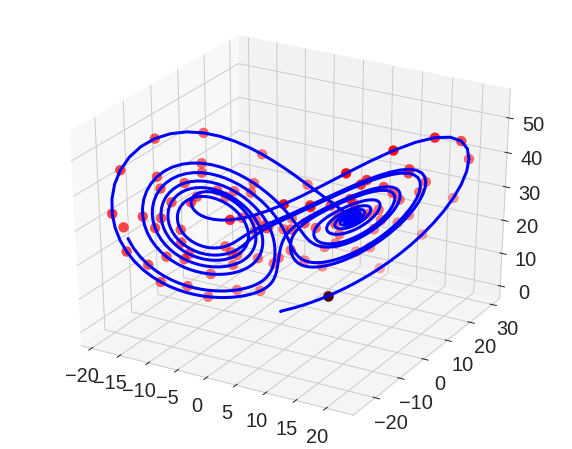}
			\caption{Lorenz Attractor}
			\label{fig:lorenz_data}
			\vspace{-3mm}
\end{wrapfigure}
As shown in Figure \ref{fig:lorenz_data}, the blue trajectory represents the discretized dynamics and red circles are sampled observations.  The dynamics is generated using $\sigma =10$ $\rho=28$, $\beta =2.667$. The initial condition of each trajectory is  sampled uniformly random from the interval of $[-0.1,0.1]$.

\begin{figure}[htbp]
\begin{center}
    \begin{subfigure}[b]{0.19\textwidth}
        \includegraphics[width=\textwidth]{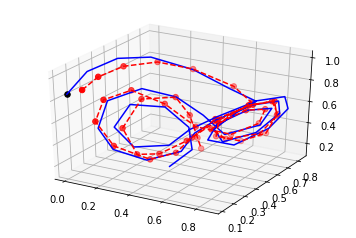}
        \caption{RNN}
    \end{subfigure}
    \begin{subfigure}[b]{0.19\textwidth}
        \includegraphics[width=\textwidth]{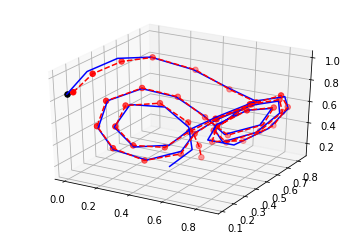}
        \caption{MRNN}
    \end{subfigure}
    \begin{subfigure}[b]{0.19\textwidth}
    \includegraphics[width=\textwidth]{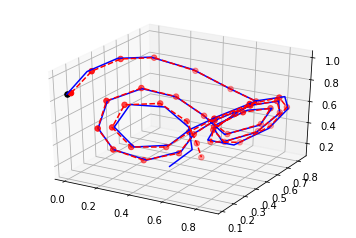}
    \caption{HORNN}
\end{subfigure}
    \begin{subfigure}[b]{0.19\textwidth}
        \includegraphics[width=\linewidth]{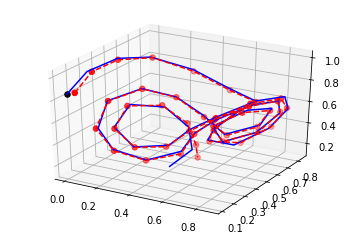}
        \caption{\trnn{}}
    \end{subfigure}
    \begin{subfigure}[b]{0.19\textwidth}
        \includegraphics[width=\linewidth]{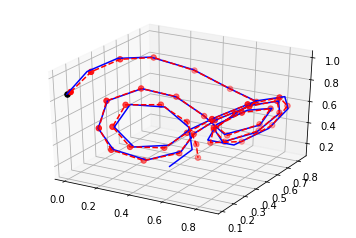}
        \caption{\tlstm{}}
    \end{subfigure} 
    \caption{Long-term (right 2) predictions for different models (red) versus the ground truth (blue). \trnn{} shows more consistent, but imperfect, predictions, whereas the baselines are highly unstable and gives noisy predictions. }
    \label{fig:long-term}
\end{center}
\vskip -0.3in
\end{figure}

Figure \ref{fig:long-term} shows $45$ steps ahead predictions for all models. HORNN is the full tensor \trnn{} using vanilla RNN unit without the tensor-train decomposition. We can see all the tensor models perform better than vanilla RNN or MRNN. \trnn{} shows slight improvement at the beginning state.

\begin{figure*}[h]
	\begin{center}
		\begin{subfigure}[b]{0.23\textwidth}		\includegraphics[width=\textwidth]{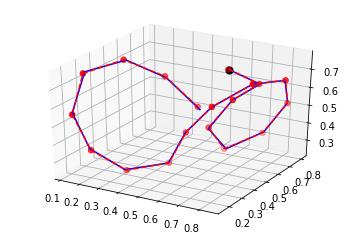}
			\caption{$T=20$}
			\label{fig:lorenz_20}
		\end{subfigure}
		\begin{subfigure}[b]{0.23\textwidth}			\includegraphics[width=\textwidth]{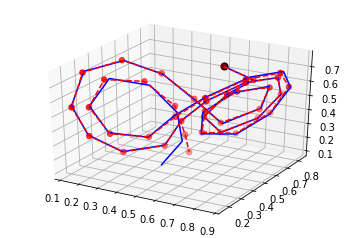}
			\caption{$T=40$}
			\label{fig:lorenz_40}
		\end{subfigure}
		\begin{subfigure}[b]{0.23\textwidth}
			\includegraphics[width=\textwidth]{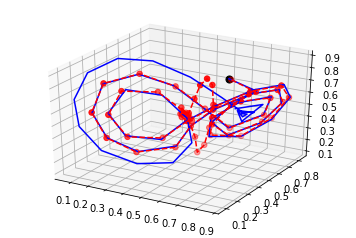}
			\caption{$T=60$}
			\label{fig:lorenz_60}
		\end{subfigure}
		\begin{subfigure}[b]{0.23\textwidth}
			\includegraphics[width=\textwidth]{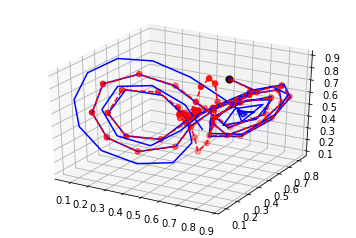}
			\caption{$T=80$}
			\label{fig:lorenz_80}
			\label{fig:f1}
		\end{subfigure}
		\caption{\ref{fig:lorenz_data} Lorenz attraction  with dynamics (blue) and sampled data (red).  \ref{fig:lorenz_20}, \ref{fig:lorenz_40}, \ref{fig:lorenz_60} ,\ref{fig:lorenz_80}   \tlstm{} long-term predictions for different forecasting horizons $T$ versus the ground truth (blue). \tlstm{} shows consistent predictions over increasing horizons $T$.}
		\label{fig:lorenz}
	\end{center}
\vskip -0.5in
\end{figure*}

We have also evaluated \trnn{} on long-term forecasting for \textit{chaotic} dynamics, such as the Lorenz dynamics. Such dynamics are highly sensitive to input perturbations: two close points can move exponentially far apart under the dynamics. This makes long-term forecasting highly challenging, as small errors can lead to catastrophic long-term errors. Figure \ref{fig:lorenz} shows that \trnn{} can predict up to $T=40$ steps into the future, but diverges quickly beyond that. We have found no state-of-the-art prediction model is stable beyond $40$ time step in this setting.

\section{Discussion}
\label{disc}
In this paper, We studied long-term forecasting  under nonlinear dynamics. We proposed a novel class of RNNs -- \trnn{} that directly learns the nonlinear dynamics using higher-order structures. We  provided the first  approximation guarantees for its representation power. We demonstrated the benefits of \trnn{} to forecast accurately for significantly longer time horizon in both synthetic and real-world multivariate time series data.

In terms of future work, forecasting \emph{chaotic dynamics}, still presents a significant challenge to any sequential prediction model. Hence, it would be worthwhile to study how to learn robust models for chaotic dynamics. For other sequence modeling tasks, such as language, there does not (or is not known to) exist a succinct analytical description of the data-generating process. It would also be interesting to go beyond forecasting and further investigate the effectiveness of \trnn{}s in such domains as well.


\acks{We would like to acknowledge support for this project
from the National Science Foundation (NSF grant IIS-9988642)
and the Multidisciplinary Research Program of the Department
of Defense (MURI N00014-00-1-0637). }


\vskip 0.2in
\newpage
\bibliography{jmlr}

\newpage

\appendix
\section*{Appendix A.}
\label{app:theorem}
\subsection{Theoretical Analysis}
We provide theoretical guarantees for the proposed \trnn{} model by analyzing a class of functions that satisfy some regularity condition. For such functions, tensor-train decomposition preserve weak differentiability and yield a compact representation.
We combine this property with neural network theory to bound the approximation error for \trnn{} with one hidden layer, in terms of:
1) the regularity of the target function $f$, 2) the dimension of the input, and 3) the tensor train rank.

In the context of \trnn{}, the target function $f(\V{x})$ with $\V{x} = \V{s} \otimes \ldots \otimes \V{s}$, is the system dynamics that describes state transitions.
Let us assume that $f(\V{x})$ is a Sobolev function: $f\in\mathcal{H}^k_\mu$, defined on the input space $\T{I}= I_1 \times I_2\times \cdots I_d $, where each $I_i$ is a set of vectors. The space $\mathcal{H}^k_\mu$ is defined as the set of  functions that have bounded derivatives up to some order $k$ and are $L_\mu$-integrable:
\begin{eqnarray}
\mathcal{H}^k_\mu =  \left\{  f  \in L^2_\mu(I):\sum_{i\leq k}\|D^{(i)}f\|^2   < +\infty \right\},
\end{eqnarray}
where $D^{(i)}f$ is the $i$-th weak derivative of $f$ and $\mu \geq 0$.\footnote{A weak derivative generalizes the derivative concept for (non)-differentiable functions and is implicitly defined as: e.g. $v\in L^1([a,b])$ is a weak derivative of $u\in L^1([a,b])$ if for all smooth $\varphi$ with $\varphi(a) = \varphi(b) = 0$: $\int_a^bu(t)\varphi'(t) = -\int_a^bv(t)\varphi(t)$.}

Any Sobolev function admits a Schmidt decomposition: $f(\cdot) = \sum_{i =0}^\infty \sqrt{\lambda_i} \gamma (\cdot)_i \otimes \phi (\cdot)_i $, where $\{\lambda \}$ are the eigenvalues and $\{\gamma\}, \{ \phi\}$ are the associated eigenfunctions.
Applying the Schmidt decomposition along $x_1$, we have 
\begin{eqnarray}
f(\V{x}) = \sum_{\alpha_1} \sqrt{\lambda_{\alpha_1}}\gamma (x_1)_{\alpha_1}\phi(x_2,\cdots, x_d)_{\alpha_1}
\end{eqnarray}

We  can apply similar Schmidt decomposition along $x_2$
\begin{eqnarray}
\sqrt{\lambda_{\alpha_1}}\phi(x_2,\cdots, x_d)_{\alpha_1} = \sum_{\alpha_2} \sqrt{\lambda_{\alpha_2}}\gamma (x_2)_{\alpha_1,\alpha_2}\phi(x_3,\cdots, x_d)_{\alpha_2}
\end{eqnarray}

Recursively performing such operation, and let $\gamma(x_d)_{\alpha_{d-1},\alpha_d} = \sqrt{\lambda_{\alpha_{d-1}}}\phi(x_d)_{\alpha_{d-1}}$ and $\T{A}^j(x_j)_{\alpha_{j-1}, \alpha_j} =\gamma(x_j)_{\alpha_{j-1}, \alpha_j}$, the target function $f \in \mathcal{H}^k_\mu$ can be decomposed as:
\begin{eqnarray}
f(\V{x}) = \sum_{\alpha_0,\cdots,\alpha_d=1}^\infty
%
\T{A}^1(x_1)_{\alpha_0\alpha_1}
\cdots
\T{A}^d(x_d)_{\alpha_{d-1}\alpha_d},
\label{app:eqn:ftt}
\end{eqnarray}
where $\{\T{A}^j(\cdot)_{\alpha_{j-1}\alpha_j} \}$ are basis functions,  
%
satisfying $\langle \T{A}^j(\cdot)_{im}, \T{A}^j (\cdot)_{in} \rangle =\delta_{mn}$.
We can truncate Eqn \ref{app:eqn:ftt} to a low dimensional subspace ($\V{r}<\infty$), and obtain the \ti{functional tensor-train (FTT)} approximation of the target function $f$:
\begin{eqnarray}
f_{TT}(\mathbf{x}) = \sum_{\alpha_0,\cdots,\alpha_d=1}^\mathbf{r}
\T{A}^1(x_1)_{\alpha_0\alpha_1}
\cdots
\T{A}^d(x_d)_{\alpha_{d-1}\alpha_d}.
\label{app:eqn:ftt}
\end{eqnarray}.

FTT approximation in Eqn \ref{app:eqn:ftt} projects the target function to a subspace with finite basis. And the approximation error can be bounded using the following Lemma:
\begin{lemma}[FTT Approximation \cite{bigoni2016spectral}]
	Let $f\in \mathcal{H}^k_\mu$ be a H\"older continuous function, defined on a bounded domain $\mathbf{I} = I_1 \times \cdots \times I_d \subset \R^d$ with exponent $\alpha > 1/2$, the FTT approximation error can be upper bounded as
	\begin{eqnarray}
	\|f- f_{TT} \|^2 \leq \|f \|^2 (d-1)\frac{(r+1)^{-(k-1)}}{(k-1)}
	\end{eqnarray}
	for $r\geq 1$ and
	\begin{eqnarray}
	\lim_{r\rightarrow \infty}\|f_{TT}-f\| ^2 =0
	\end{eqnarray}
	for $k>1$
	\label{app:lemma:ftt}
\end{lemma}
Lemma \ref{app:lemma:ftt} relates  the approximation error to the  dimension  $d$,  tensor-train rank $r$,and  the regularity of the target function $k$. In practice, \trnn{} implements a polynomial expansion of the input states $\V{s}$, using powers $[\V{s}, \V{s}^{\otimes 2}, \cdots, \V{s}^{\otimes p}]$ to approximate $f_{TT}$, where $p$ is the degree of the polynomial.  We can further use the classic spectral  approximation  theory to connect the \trnn{} structure  with the degree of the polynomial, i.e., the order of the tensor. Let $I_1 \times \cdots \times I_d = \V{I} \subset \R^d $. Given a  function $f$  and its polynomial  expansion  $P_{TT}$, the approximation error  is therefore bounded by:
%
\begin{lemma}[Polynomial Approximation]
	Let $f\in  \mathcal{H}^k_\mu$ for $k>0$. Let $P$ be  the approximating polynomial with degree $p$, Then
	\[ \|f - P_N f\| \leq C(k)p^{-k}|f|_{k,\mu}  \]
\end{lemma}

Here $|f|^2_{k,\mu} =\sum_{|i|=k}\|D^{(i)}f\|^2 $ is the semi-norm of the  space $\mathcal{H}^k_\mu$.  $C(k)$ is the  coefficient of the spectral  expansion.  By definition, $\mathcal{H}^k_\mu$ is equipped with a norm $\|f\|^2_{k,\mu} =\sum_{|i|\leq k}\|D^{(i)}f\|^2$ and a semi-norm $|f|^2_{k,\mu} =\sum_{|i|=k}\|D^{(i)}f\|^2 $. For notation simplicity, we muted the subscript $\mu$ and used $\|\cdot\| $ for $\|\cdot \|_{L_{\mu}}$.

So far, we have obtained the  tensor-train approximation error with the regularity of the target function $f$. Next we will  connect the tensor-train approximation and the approximation error  of neural networks with one layer hidden units. Given a neural network with one hidden layer and sigmoid activation function, following Lemma describes the classic result of  describes the error between a target function $f$ and the single hidden-layer neural network that approximates it best:
\begin{lemma}[NN Approximation \cite{barron1993universal}]
	Given a function $f$ with finite Fourier magnitude distribution $C_f$, there exists a  neural network with $n$ hidden units $f_n$, such that
	\begin{eqnarray}
	\| f - f_n\| \leq \frac{C_f}{\sqrt{n}}
	\end{eqnarray}
	where $C_f = \int |\omega|_1  | \hat{f}(\omega) | d \omega$ with Fourier representation $f(x)=\int e^{i\omega x}\hat{f}(\omega) d\omega$.
	\label{app:lemma:nn}
\end{lemma}

We can now generalize Barron's approximation lemma \ref{app:lemma:nn} to \trnn{}.  The target time series is  $f(\V{x}) = f(\V{s}\otimes \dots \otimes \V{s})$.  We can express the function using FTT, followed by the polynomial expansion of the  states concatenation $P_{TT}$. The approximation error of \trnn{}, viewed as one layer hidden 

\begin{eqnarray*}
	\|f- P_{TT}\| & \leq& \|f- f_{TT}\| +\| f_{TT} - P_{TT}\| \\
	&\leq& \|f\|\sqrt{ (d-1)\frac{(r+1)^{-(k-1)}}{(k-1)}} + C(k)p^{-k}|f_{TT}|_k\\
	&\leq& \|f-f_n\|\sqrt{ (d-1)\frac{(r+1)^{-(k-1)}}{(k-1)}} + C(k)p^{-k} \sum\limits_{i=k}  \|D^{(i)}(f_{TT}-f_n)\|  +o(\|f_n\|)\\
	&\leq & \frac{C^2_f}{n} (\sqrt{ (d-1)\frac{(r+1)^{-(k-1)}}{(k-1)}} + C(k)p^{-k} \sum\limits_{i=k}  \|D^{(i)}f_{TT}\| ) +o(\|f_n\|)
\end{eqnarray*}

Where $p$ is the order of tensor and $r$ is the tensor-train rank. As the rank of the tensor-train and the polynomial order increase, the required size of the hidden units become smaller, up to a constant that depends on the regularity of the underlying dynamics $f$.

\subsection{Additional Experiments}
\begin{figure}[htbp]
\begin{center}
        \begin{subfigure}[b]{\textwidth}
            \includegraphics[width=\textwidth]{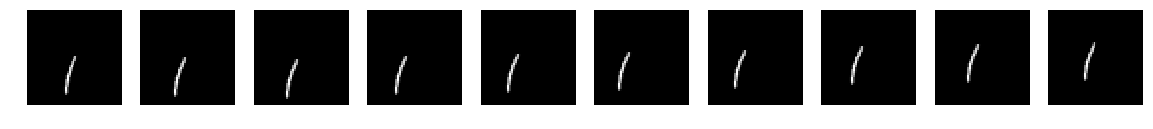}
            \label{app:fig:lorenz_80}
            \label{app:fig:f1}
        \end{subfigure}\\
                \begin{subfigure}[b]{\textwidth}
            \includegraphics[width=\textwidth]{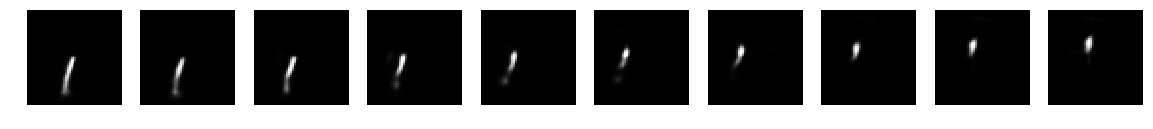}
            \label{app:fig:lorenz_80}
            \label{app:fig:f1}
        \end{subfigure}\\
                \begin{subfigure}[b]{\textwidth}
            \includegraphics[width=\textwidth]{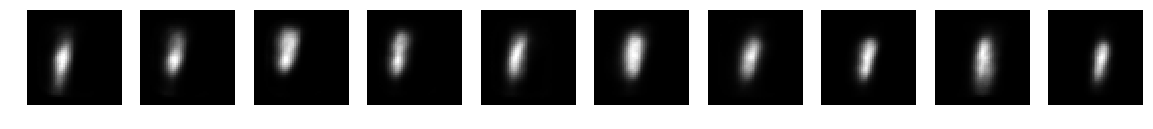}
            \label{app:fig:lorenz_80}
            \label{app:fig:f1}
        \end{subfigure}
         \caption{Visualizations of  ground truth and predictions from \tlstm{} and baselines for moving MNIST. Top: ground truth; Middle: LSTM predictions; Bottom: \tlstm{} predictions.}
    \label{app:fig:genz}
     \vskip -0.2in
    \end{center}
\end{figure}

\paragraph{Genz dynamics}
Genz functions  are often used as basis for evaluating high-dimensional function approximation.  Figure \ref{app:fig:genz} visualizes different Genz  functions, realizations of dynamics and predictions from \tlstm{} and baselines.  We can see for ``oscillatory'', ``product peak'' and ``Gaussian '', \tlstm{} can better capture the complex dynamics, leading to more accurate predictions.

\begin{figure}[htbp]
    \centering
    \begin{subfigure}[b]{0.3\textwidth}
        \includegraphics[width=\textwidth]{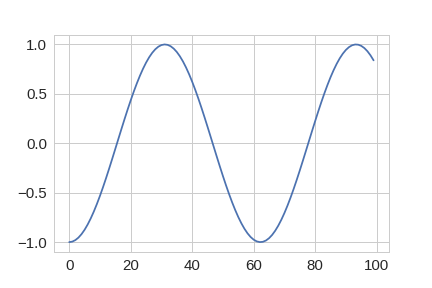}
        \caption{$g_1$ oscillatory}
        \label{app:fig:f1}
    \end{subfigure}
    ~ 
    \begin{subfigure}[b]{0.3\textwidth}
        \includegraphics[width=\textwidth]{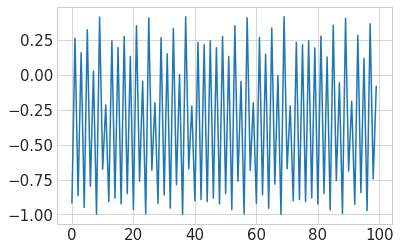}
        \caption{$g_1$ dynamics}
        \label{app:fig:df1}
    \end{subfigure}
    \begin{subfigure}[b]{0.3\textwidth}
        \includegraphics[width=\textwidth]{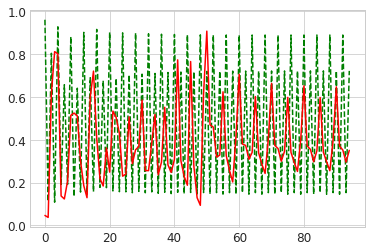}
        \caption{$g_1$ predictions}
        \label{app:fig:f2}
    \end{subfigure}\\ 

        \begin{subfigure}[b]{0.3\textwidth}
        \includegraphics[width=\textwidth]{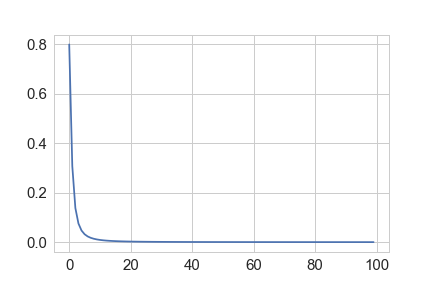}
        \caption{$g_2$ product peak}
        \label{app:fig:f2}
    \end{subfigure}
    ~ 
    \begin{subfigure}[b]{0.3\textwidth}
        \includegraphics[width=\textwidth]{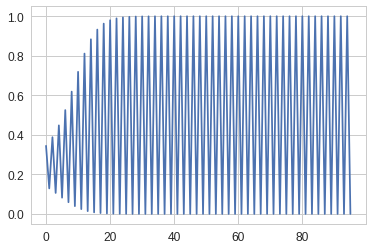}
        \caption{$g_2$ dynamics}
        \label{app:fig:df2}
    \end{subfigure}
    ~ 
    \begin{subfigure}[b]{0.3\textwidth}
        \includegraphics[width=\textwidth]{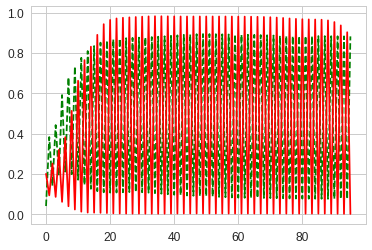}
        \caption{$g_2$ predictions}
        \label{app:fig:df2}
    \end{subfigure}\\ 
            \begin{subfigure}[b]{0.3\textwidth}
        \includegraphics[width=\textwidth]{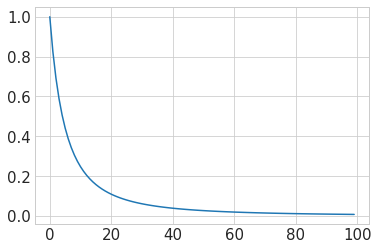}
        \caption{$g_3$ corner peak}
        \label{app:fig:f3}
    \end{subfigure}
    ~ 
    \begin{subfigure}[b]{0.3\textwidth}
        \includegraphics[width=\textwidth]{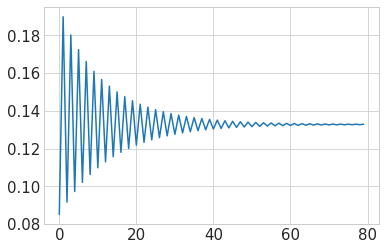}
        \caption{$g_3$ dynamics}
        \label{app:fig:df3}
    \end{subfigure}
    ~ 
    \begin{subfigure}[b]{0.3\textwidth}
        \includegraphics[width=\textwidth]{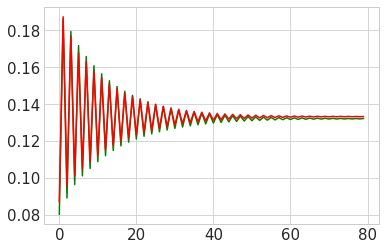}
        \caption{$g_3$ predictions}
        \label{app:fig:df3}
    \end{subfigure}\\ 

    \begin{subfigure}[b]{0.3\textwidth}
        \includegraphics[width=\textwidth]{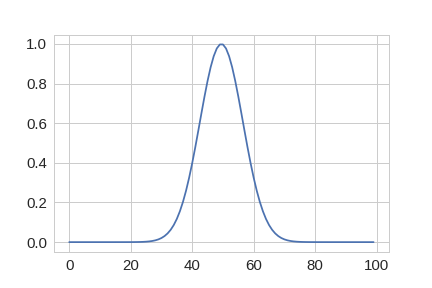}
        \caption{$g_4$ Gaussian}
        \label{app:fig:f4}
    \end{subfigure}
    \begin{subfigure}[b]{0.3\textwidth}
        \includegraphics[width=\textwidth]{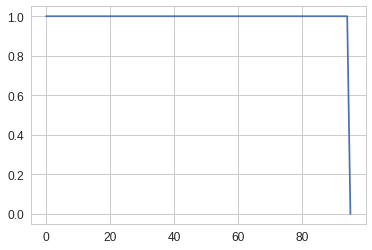}
        \caption{$g_4$ dynamics}
        \label{app:fig:df4}
    \end{subfigure}
    \begin{subfigure}[b]{0.3\textwidth}
        \includegraphics[width=\textwidth]{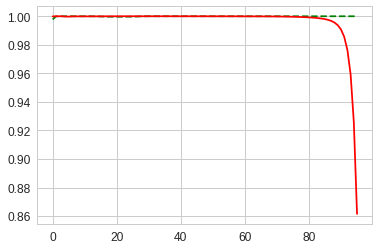}
        \caption{$g_4$ predictions}
        \label{app:fig:f4}
    \end{subfigure}\\ 

        \begin{subfigure}[b]{0.3\textwidth}
        \includegraphics[width=\textwidth]{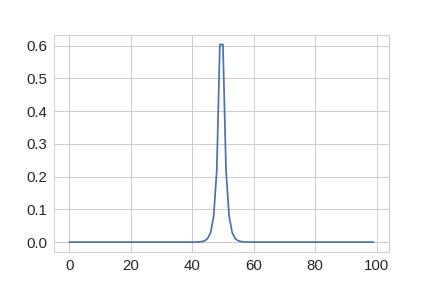}
        \caption{$g_5$ continuous}
        \label{app:fig:f5}
    \end{subfigure}
    ~ 
    \begin{subfigure}[b]{0.3\textwidth}
        \includegraphics[width=\textwidth]{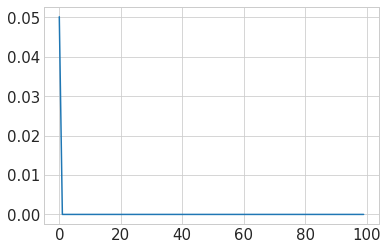}
        \caption{$g_5$ dynamics}
        \label{app:fig:df5}
    \end{subfigure}
    ~ 
    \begin{subfigure}[b]{0.3\textwidth}
        \includegraphics[width=\textwidth]{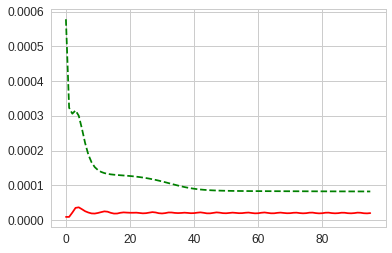}
        \caption{$g_5$ predictions}
        \label{app:fig:df5}
    \end{subfigure}\\ 

            \begin{subfigure}[b]{0.3\textwidth}
        \includegraphics[width=\textwidth]{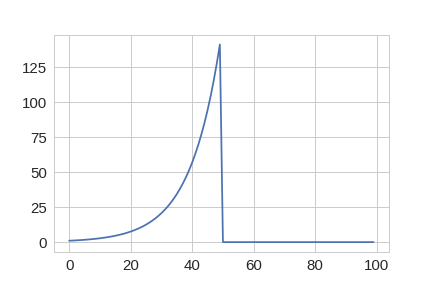}
        \caption{$g_6$ discontinuous}
        \label{app:fig:f6}
    \end{subfigure}
    ~ 
    \begin{subfigure}[b]{0.3\textwidth}
        \includegraphics[width=\textwidth]{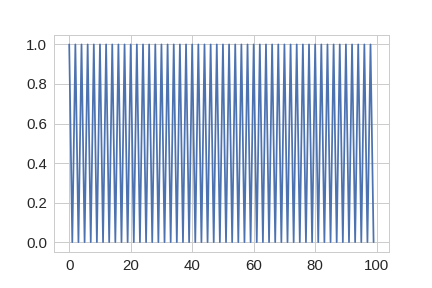}
        \caption{$g_6$ dynamics}
        \label{app:fig:df6}
    \end{subfigure}
    ~ 
    \begin{subfigure}[b]{0.26\textwidth}
        \includegraphics[width=\textwidth]{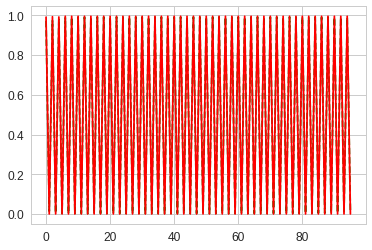}
        \caption{$g_6$ predictions}
        \label{app:fig:f6}
    \end{subfigure}
    \caption{Visualizations of  Genz  functions,  dynamics and predictions from \tlstm{} and baselines. Left column: transition functions, middle: realization of the dynamics and right: model predictions for LSTM (green) and \tlstm{} (red).}
    \label{fig:genz}
 \vskip -0.2in
 \end{figure}

\paragraph{Moving MNIST}
 Moving MNIST \cite{srivastava2015unsupervised} generates around $50,000$ video sequences of length $100$ on the fly. The video is generated by moving the digits in the MNIST image dataset along a given trajectory  within a canvas of size $48\times 48$. The trajectory reflects the dynamics of the movement. In this experiment, we used $cos$ and $sin$ velocity.

\end{document}